\documentclass[sigconf]{acmart}

\usepackage{booktabs, tabularx, threeparttable, algorithm, amsmath, amsfonts, mathtools, multirow, multicol, makecell}
\usepackage{xcolor}
\usepackage{footmisc}

\usepackage[noend]{algpseudocode}

\algrenewcommand\textproc{}
\makeatletter
\newcommand{\multiline}[1]{%
	\begin{tabularx}{\dimexpr\linewidth-\ALG@thistlm}[t]{@{}X@{}}
		#1
	\end{tabularx}
}

\algnewcommand{\Initialize}[1]{%
  \State \textbf{Initialize:} \parbox[t]{.8\linewidth}{\raggedright #1}
}
\algnewcommand\algorithmicforeach{\textbf{for each}}
\algdef{S}[FOR]{ForEach}[1]{\algorithmicforeach\ #1\ \algorithmicdo}

\AtBeginDocument{%
  }

\begin{document}

\title{A Lightweight, Efficient and Explainable-by-Design Convolutional Neural Network for Internet Traffic Classification}

\author{Kevin Fauvel}
\affiliation{
  \institution{Huawei Technologies Co., Ltd}
  \country{France}
}

\author{Fuxing Chen}
\affiliation{
	\institution{Huawei Technologies Co., Ltd}
	\country{France}
}

\author{Dario Rossi}
\affiliation{
	\institution{Huawei Technologies Co., Ltd}
	\country{France}
}

\renewcommand{\shortauthors}{Fauvel et al.}
\newcommand{\ournet}{LEXNet}

\begin{abstract}
Traffic classification, i.e., the identification of the type of applications flowing in a network, is a strategic task for numerous activities (e.g., intrusion detection, routing). This task faces some critical challenges that current deep learning approaches do not address. The design of current approaches do not take into consideration the fact that networking hardware (e.g., routers) often runs with limited computational resources. Further, they do not meet the need for faithful explainability highlighted by regulatory bodies. Finally, these traffic classifiers are evaluated on small datasets which fail to reflect the diversity of applications in real-world settings.

Therefore, this paper introduces a new Lightweight, Efficient and eXplainable-by-design convolutional neural network (\ournet) for Internet traffic classification, which relies on a new residual block (for lightweight and efficiency purposes) and prototype layer (for explainability). Based on a commercial-grade dataset, our evaluation shows that \ournet\ succeeds to maintain the same accuracy as the best performing state-of-the-art neural network, while providing the additional features previously mentioned. Moreover, we illustrate the explainability feature of our approach, which stems from the communication of detected application prototypes to the end-user, and we highlight the faithfulness of \ournet\ explanations through a comparison with post hoc methods.
\end{abstract}



\maketitle

\section{Introduction}
Traffic classification, i.e., the identification of the type of applications flowing in a network~\citep{Dainotti12}, is strategic for network operators to perform a wide range of network management activities~\cite{Boutaba18, Aceto20}. These activities include capacity planning, intrusion detection and service differentiation (e.g., for business critical applications).

The sharp rise of encrypted traffic in the last decade hampered traditional rule-based techniques, pushing the adoption of machine learning-assisted classification~\citep{pacheco18comst} to supplant deep packet inspection~\citep{ntop14, Cisco18, RohdeSchwarz20, Huawei21}. Inspired by seminal work~\cite{crotti07ccr}, a recent wave~\citep{Liu19, Aceto19mimetic, Aceto20, Beliard20, Lotfollahi20, Rezaei20, Wang20, Nascita21} of deep learning models (Convolutional Neural Networks - CNNs, or recurrent architectures like Long Short-Term Memory - LSTMs) successfully tackles the encryption challenge by just leveraging the size and direction of the first few packets in a flow.

However, these studies still face major challenges.
First, due to the widespread use of mobile devices and the vast diversity of mobile applications, \emph{large-scale} encrypted traffic classification becomes increasingly difficult~\cite{Aceto19, Aceto20}. At the same time, academic models are generally evaluated with a few tens of classes, which are significantly below commercial settings (hundreds to thousands classes)~\cite{Cisco18, RohdeSchwarz20, Huawei21}.
Second, even though focusing on relatively small datasets, academic models may still use a disproportionate amount of resources - models with millions of weights are commonplace as highlighted in~\cite{Yang21}. This clashes with the need for near real-time classification (e.g., due to latency sensitive traffic) on the one hand, and the  \emph{limited computational resources}~\cite{Dias19} available on network devices (e.g., routers) on the other, which renders a lightweight and efficient model necessary. Indeed, fast inference ($\sim$10k classifications/second) on traffic classification engines having limited computational budget (e.g., ARM CPUs, microcontrollers) requires model to be small in order to experience fewer computation during forward propagation.
Last but not least, regulatory and standardization bodies highlight \emph{faithful explainability} as a pillar for accountability, responsibility, and transparency of processes including AI components~\cite{Dignum17, EU21, Phillips21}, which could prevent the deployment of the latest machine learning techniques that do not have this feature.
Faithfulness is critical as it corresponds to the level of trust an end-user can have in the explanations of model predictions, i.e., the level of relatedness of the explanations to what the model actually computes~\cite{Fauvel20_IJCAI}. In the context of traffic classification, a faithfully explainable model (explainable-by-design) would be beneficial from both a compliance and a business standpoint (e.g., SLAs). Nonetheless, the faithfulness of such a model should not negatively impact the prediction performance, nor come at the cost of an increased model computational and memory complexity. A few models in traffic classification~\cite{Beliard20, Nascita21, Rezaei20} account for explainability, but they cannot provide faithful explainability as they rely on post hoc model-agnostic explainability~\cite{Rudin19}.
With the above constraints in mind, we therefore propose a new Lightweight, Efficient and eXplainable-by-design CNN (\ournet) for traffic classification, and evaluate it on a large-scale commercial-grade dataset. 

\begin{figure*}[!htpb]
	\centering
	\includegraphics[width=0.76\linewidth]{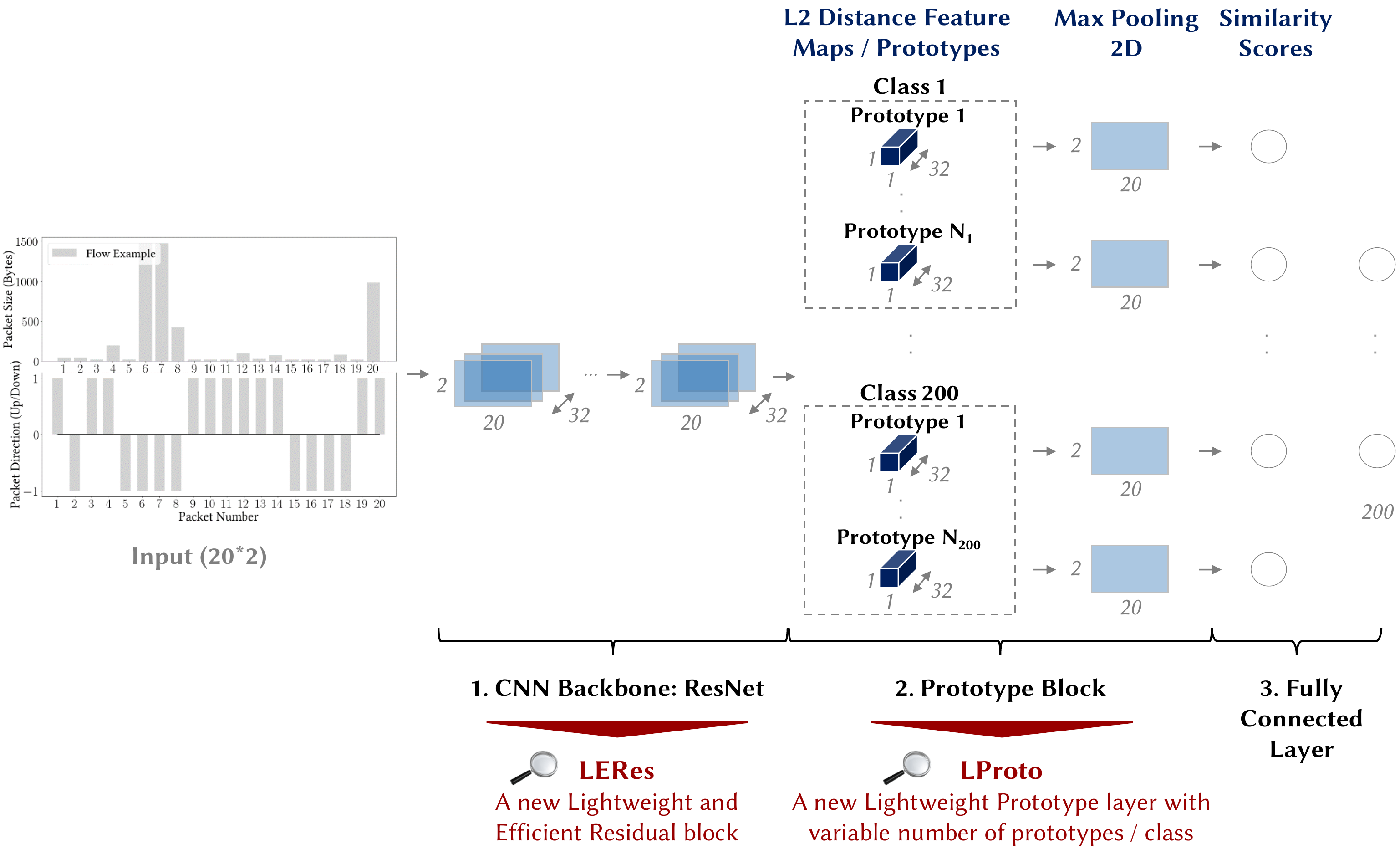}
	\caption{Illustration of LEXNet 3-part structure, on our commercial-grade dataset composed of 200 classes, with its two main contributions in red: LERes block and LProto layer. LERes is detailed in Figure \ref{fig:LERes} and LProto in Figure \ref{fig:LProto}. N - Number.}
	\label{fig:LEXNet}
\end{figure*}

An additional requirement that shapes the overall architecture of our classifier is the form of explanations. Based on network experts input, we have adopted explanations under the form of class-specific prototypes as they can be associated with application signatures and provided for each sample/flow to the end-user. Thus, as illustrated in Figure~\ref{fig:LEXNet}, the overall architecture of our \ournet\ follows the typical 3-part structure of explainable-by-design prototype-based networks~\cite{Chen19}: 
\textit{(i)} a CNN backbone extracts discriminative features, then \textit{(ii)} a prototype block computes similarities to the learned class-specific prototypes, and finally \textit{(iii)} the classification is performed based on these similarity scores.
Then, the contributions of this paper are the following:

\begin{itemize}
    \item \textit{(i) CNN backbone: LERes block} - \ournet\ redesigns the widely used residual block~\cite{He16} with cost-efficient operations (linear transformations and concatenate operations) which significantly reduce its number of parameters (-19\%) and CPU inference time (-41\%), while preserving the accuracy of the network (-0.7\%). We called this new block LERes block;
    \item \textit{(ii) Prototype block: LProto} - \ournet\ converts the current state-of-the-art prototype block~\cite{Chen19} into a unique prototype layer (called LProto). Preserving the explainability-by-design, LProto significantly reduces the number of parameters (-36\%) and CPU inference time (-24\%), while increasing the accuracy by 4\%. In addition, in contrast with the prototype block which uses a unique number of prototypes per class as hyperparameter - same number for all classes, LProto automatically learns a variable number of prototypes per class during training to better characterize the discriminative features of each application;
    \item \textit{Evaluation: Performance} - The evaluation on our commercial-grade dataset shows that \ournet\ is more accurate than the current state-of-the-art explainable-by-design CNN (ProtoPNet~\cite{Chen19}), and maintains the same accuracy as the best performing deep learning approach (ResNet~\cite{He16}) that does not provide faithful explanations. Moreover, we demonstrate that \ournet\ significantly reduces the model size and inference time compared to the state-of-the-art neural networks with explainability-by-design and post hoc explainability methods. 
    Both our dataset~\cite{figshare} and code~\cite{github} are made publicly available to allow reproducibility. In addition, we evaluate \ournet\ on current public benchmark datasets from different domains and show that the results are consistent with the ones obtained on our dataset;
    \item \textit{Evaluation: Explainability} - Following the illustration of the explainability of our approach which stems from the communication of detected application/class prototypes to the end-user, we highlight the faithfulness of \ournet\ explanations by comparing them to the ones provided by the state-of-the-art post hoc explainability methods.
\end{itemize}

\section{Related Work}
\label{sec:related_work}
Traffic classification can be formulated as a Multivariate Time Series (MTS) classification task, where the input is packet-level data (typically packet size and direction) related to a single flow\footnote{A packet is a unit of data routed on the network, and a flow is a set of packets that shares the same 5-tuple (i.e., source IP, source port, destination IP, destination port, and transport-layer protocol)~\cite{Valenti13}.}, and the output is an application label.
An example of a flow is represented in grey in Figure~\ref{fig:LEXNet}, with the packet size and direction (y-axis) of the first 20 packets (x-axis) of a flow belonging to Application 1. 
This paper addresses encrypted traffic classification as a MTS problem, when it has been tackled as an univariate one in the literature ($\pm$ packet size as input time series). Our multivariate approach offers more granularity and information in its explanations, with both per-packet size and direction as explanatory variables, without degradation in performance.

The state-of-the-art approaches~\cite{Aceto19mimetic, Beliard20, Liu19, Lotfollahi20, Nascita21, Rezaei20, Wang20} adopt heavyweight deep learning classifiers (from 1M to 2M weights), and do not discuss the impact of their model size on the accuracy and inference time.
When accounting for explainability, these studies employ post hoc explainability methods like SHAP~\cite{Lundberg17}, which cannot meet the faithfulness requirement. Finally, the results of these studies are not directly comparable as they are evaluated on different datasets. Plus, these datasets (mostly private, from 8 to 80 applications and from 8k to 950k flows) fail to reflect the diversity of applications in real commercial settings.

Consequently, an accurate, lightweight, efficient and explainable-by-design approach for traffic classification evaluated on a public commercial-grade dataset is necessary. We identified CNNs as having the potential to fulfill these needs. 
Despite the recent promising results of large multimodal models and their potential for addressing various network management tasks, we do not consider them in this study as these large models are not compatible with the limited computational hardware resources available in networks today, nor are they capable of integrating explainability-by-design for the extraction of faithful explanations to support their predictions.
In the next sections, we present the corresponding machine learning state-of-the-art (MTS classification, explainability, lightweight and efficient architectures) on which we position our approach.

\subsection{MTS Classification}
\label{sec:related:mts}
The state-of-the-art MTS classifiers are composed of three categories: similarity-based~\cite{Seto15}, feature-based~\cite{Baydogan16, Karlsson16, Schafer17, Tuncel18} and deep learning methods~\cite{Karim19, Zhang20, Fauvel21}. The results published show that the top two most accurate MTS classifiers on average on the public UEA archive~\cite{Bagnall18} are deep learning methods (top-1 XCM~\cite{Fauvel21}, top-2 MLSTM-FCN~\cite{Karim19}). XCM extracts discriminative features related to observed variables and time directly from the input data using 2D and 1D convolutions filters, while MLSTM-FCN stacks a LSTM layer and a 1D CNN layer along with squeeze-and-excitation blocks to generate latent features.
Therefore, in this work, we choose to benchmark \ournet\ to these two MTS classifiers. Plus, we integrate in our benchmark the commonly adopted state-of-the-art CNNs in computer vision (ResNet~\cite{He16}, DenseNet~\cite{Huang17}), the vanilla 1D CNN widely used in traffic classification studies, and the traditional machine learning methods Random Forest~\cite{Breiman01} and XGBoost~\cite{Chen16}.

\subsection{Explainability}
\label{sec:related:xai}
There are several methods belonging to two main categories~\cite{Du20}: explainability-by-design and post hoc explainability. 

Post hoc methods are the most popular ones for deep learning models (model-specific methods as the saliency method Grad-CAM~\cite{Selvaraju17}, or model-agnostic as the surrogate model SHAP~\cite{Lundberg17}). However, some recent studies~\cite{Rudin19, Chen20} show that post hoc explainability methods face a faithfulness issue.

Lately, there have been some work proposing explainable-by-design CNN approaches~\cite{Zhang18, Chen19, Elsayed19, Chen20} to circumvent this issue. 
Two of these approaches~\cite{Zhang18, Elsayed19} provide the explainability-by-design feature but at the cost of some lag behind the state-of-the-art CNNs in terms of accuracy, which remains a prerequisite for our task.
Then, Chen et al. (2020) present Concept Whitening, an alternative to a batch normalization layer which decorrelates the latent space and aligns its axes with known concepts of interest. This approach requires that the concepts/applications are completely decorrelated, so it is not suited for our traffic classification task as some applications can be correlated (e.g., different applications from the same editor).
Finally, another approach combines accuracy with explainability-by-design: ProtoPNet~\cite{Chen19}.
It consists of a CNN backbone to extract discriminative features, then a prototype block that computes similarities to the learned class-specific prototypes as basis for classification.
This approach is particularly interesting for our task as it is aligned with the way humans describe their own thinking in classification, by focusing on parts of the input data and comparing them with prototypical parts of the data from a given class. In particular, prototypes can be used to form application signatures for network experts.
Therefore, with the objective to combine accuracy and explainability-by-design suited for traffic classification, we adopted a prototype-based CNN approach. Then, we worked to reduce the number of weights and inference time of the approach while maintaining the accuracy, in particular with regard to the CNN backbone.

\subsection{Lightweight and Efficient Architectures}\label{sec:related:cost}
Multiple studies about lightweight and efficient CNN architectures have been published over the last years: ShuffleNetV2~\cite{Ma18} and its four practical guidelines, EfficientNet~\cite{Tan19} with its scaling method along depth/width/resolution dimensions, MobileNetV3~\cite{Howard19} with its network search techniques and optimized nonlinearities, GhostNet~\cite{Han20} using a Ghost module based on linear transformations, and CondenseNetV2~\cite{Yang21CN2} relying on a Sparse Feature Reactivation module. 

These CNNs are evaluated on the same public dataset (ImageNet).
However, most of these evaluations optimize a proxy metric to measure efficiency (number of floating-point operations per second - FLOPs), which is not always equivalent to the direct metric we care about - the inference time~\cite{Ma18}. Using FLOPs as the reference, these studies do not evaluate the accuracy and inference time on a comparable model size (number of weights).
A recent work~\cite{Ridnik21} shows that ResNet architecture is usually faster than its latest competitors, offering a better accuracy/inference time trade-off (which is also confirmed by our experiments in Section~\ref{sec:classification}).
Therefore, we adopted ResNet as CNN backbone and revisited its residual block to make it lighter and more efficient. We include ShuffleNetV2, EfficientNet, MobileNetV3, GhostNet and CondenseNetV2 in our benchmark, and perform a comparison of the accuracy and inference time based on a comparable model size in Section~\ref{sec:classification}.

\section{Algorithm}
\label{sec:lexnet}
We now propose our new Lightweight, Efficient and eXplainable-by-design CNN classifier (\ournet). 
The overall architecture of \ournet\ follows the one of explainable-by-design prototype-based networks like the state-of-the-art ProtoPNet\cite{Chen19}.
Before introducing our contributions, we present the structure and functioning of a prototype-based network.
As illustrated in Figure~\ref{fig:LEXNet}, the network consists of three parts. 
First, it is composed of a CNN backbone for feature extraction (see 1. in Figure~\ref{fig:LEXNet}). In the context of \ournet, as shown in section~\ref{sec:classification}, ResNet~\cite{He16} has been selected as it obtains the best accuracy.
Then, a prototype block learns class-specific prototypes, with the size of the prototypes set as hyperparameter of the algorithm. Prototypes of size (1, 1) are chosen (see 2. in Figure~\ref{fig:LEXNet}) to be able to precisely identify discriminative features and provide valuable explanations to the end-user.
Each prototype in the block computes the L2 distances with all patches of the last convolutional layer (e.g., size of the last layer in Figure~\ref{fig:LEXNet}: $20 \times 2 \times 32$), and inverts the distances into similarity scores.
The result is an activation map of similarity scores (e.g., size of a similarity matrix in Figure~\ref{fig:LEXNet}: $20 \times 2$). 
This activation map preserves the spatial relation of the convolutional output, and can be upsampled to the size of the input sample to produce a heatmap that identifies the part of the input sample most similar to the learned prototype.
To make sure that the learned class-specific prototypes correspond to training samples patches, the prototypes are projected onto the nearest latent training patch from the same class during the training of the model.
The activation map of similarity scores produced by each prototype is then reduced using global max pooling to a single similarity score, which indicates how strong the presence of a prototypical part is in some patch of the input sample.
Finally, the classification is performed with a fully connected layer based on these similarity scores (see 3. in Figure~\ref{fig:LEXNet}). 

Then, \ournet\ has two main contributions.
Our first contribution is to introduce a new lightweight and efficient residual block which reduces the number of weights and inference time of the original residual block, while preserving the accuracy of ResNet.
To further reduce the size of \ournet\ and improve its accuracy, we propose as a second contribution to replace the prototype block with a new lightweight prototype layer. This new prototype layer also allows a better characterization of each class by automatically learning a variable number of prototypes per class.
We present in the next sections these two contributions, and end with the overall architecture of our network and its training procedure.

\begin{figure}[!htpb]
	\centering
	\includegraphics[width=\linewidth]{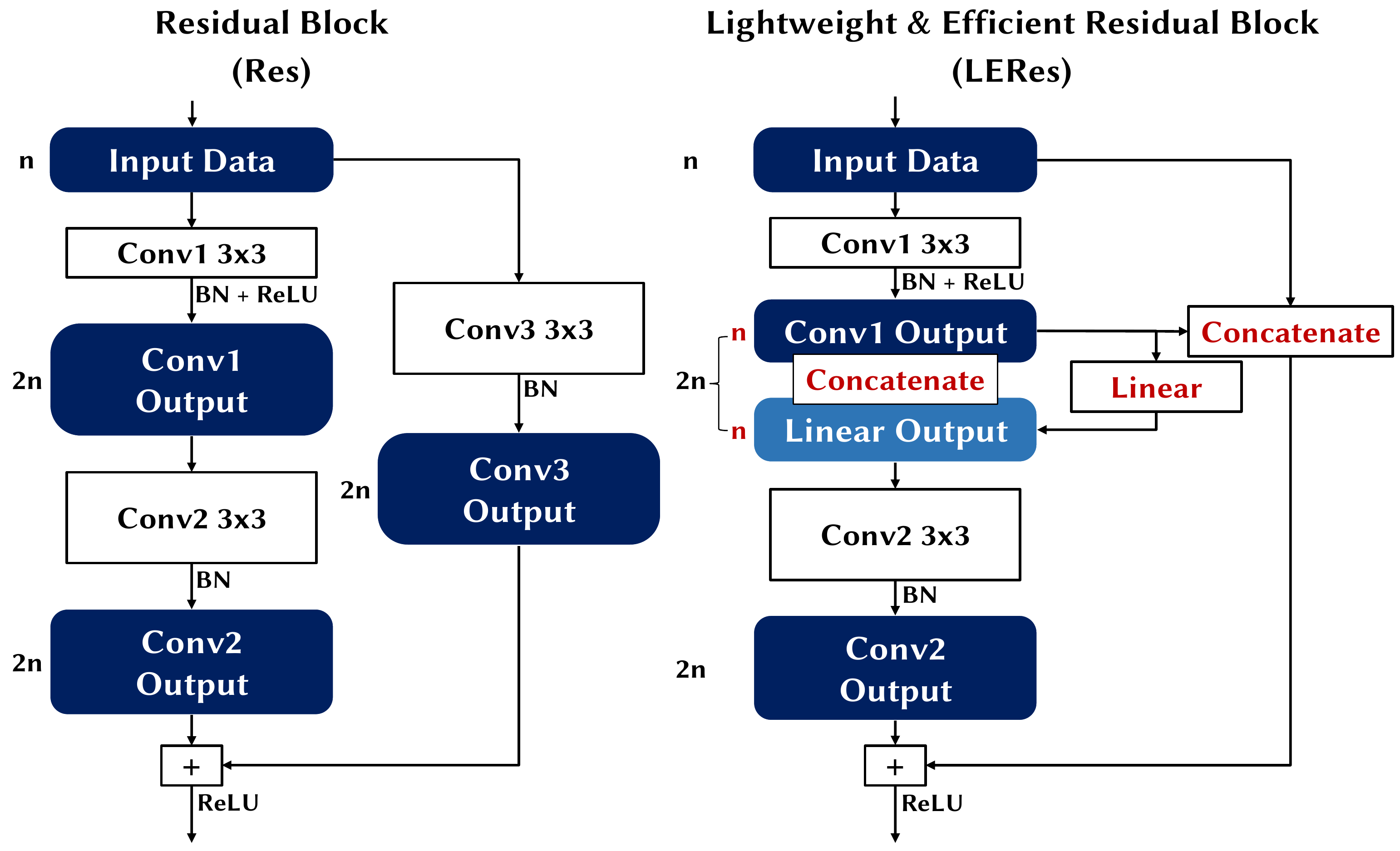}
	\caption{ResNet original residual block (Res) and our new Lightweight and Efficient Residual block (LERes) with its contributions in red.}
	\label{fig:LERes}
\end{figure}

\subsection{Lightweight and Efficient Residual Block}
\label{sec:LERes}
As previously introduced, we select ResNet~\cite{He16} for the CNN backbone as it obtains the best accuracy on our traffic classification dataset (detailed in Section~\ref{sec:classification}).
ResNet is a state-of-the-art CNN composed of consecutive residual blocks which consist of two convolutions with a shortcut added to link the output of a block to its input in order to reduce the vanishing gradient effect~\cite{Zagoruyko16}. 
When the number of output channels equals the one of the input channels, the input data is directly added to the output of the block with the shortcut.
Nonetheless, the prediction performance of ResNet is supported by an increase in the number of channels along the depth of the network to learn elaborated features, which implies residual blocks with a higher number (usually twice) of output channels than input channels.
In this case, before performing the addition, another convolution is applied to the input data in order to match the dimensions of the block output. A residual block is illustrated on the left side of Figure~\ref{fig:LERes} with n input channels and 2n output channels. We can see in this Figure that two convolutions have a different number of output channels than input channels (Convolutions 1 and 3).
However, a guideline from~\cite{Ma18} (ShuffleNetV2) states that efficient convolutions should keep an equal channel width to minimize memory access cost and inference time.
Therefore, we propose a new residual block that only performs convolutions with an equal channel width, and the additional feature maps are generated with two cheap operations (see right side of Figure~\ref{fig:LERes}).

First, the authors of GhostNet~\cite{Han20} state that CNNs with good prediction performance have redundancy in feature maps, and that substituting some of the feature maps with a linear transformation of the same convolution output does not impact the prediction performance of the network. Therefore, we double the number of feature maps from the first convolution in a cost-efficient way using a series of linear transformations of its output ($3 \times 3$ linear kernel). These new feature maps are concatenated to the ones from the first convolution output to form the input of the second convolution\footnote{We have also considered having both convolutions 1 and 2 with an equal channel width n, and moving the concatenate operation with the feature maps from the linear transformation after the  convolution 2. However, our experiments showed that this configuration leads to a decrease in accuracy without noticeable drop in inference time.}. Thus, the first and second convolutions in our new residual block have an equal channel width (n and 2n respectively, see Figure~\ref{fig:LERes}).

The second operation concerns the shortcut link: instead of converting the input data to the block output dimensions with a third convolution (see Res block in Figure~\ref{fig:LERes}), we propose to save this convolution by keeping the input data and concatenating it with the output from the first convolution.
Our experiments on the traffic classification dataset show that the new residual block LERes allows ResNet to reduce the number of weights of the backbone by 19.1\% and the CPU inference time by 41.3\% compared to ResNet with the original residual block, while maintaining its accuracy (99.3\% of the original accuracy).

\begin{figure}[!htpb]
	\centering
	\includegraphics[width=\linewidth]{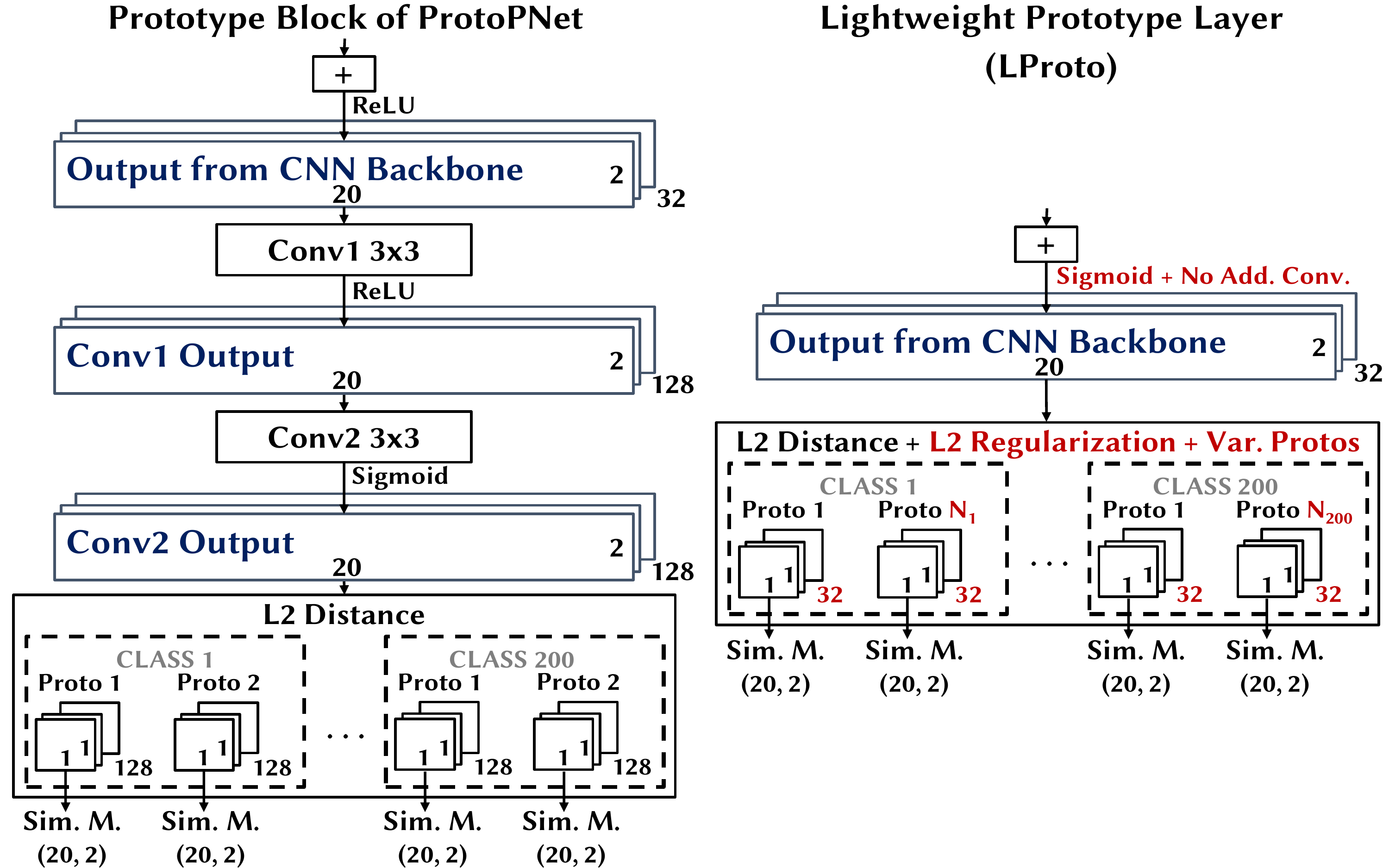}
	\caption{The prototype block of ProtoPNet and the new Lightweight Prototype Layer of \ournet\ (LProto) with its contributions in red. Sim. M. - similarity matrix.}
	\label{fig:LProto}
\end{figure}

\subsection{Lightweight Prototype Layer}
Next, we adopt a new lightweight prototype layer instead of the prototype block of ProtoPNet. As illustrated in Figure~\ref{fig:LProto}, the prototype block of ProtoPNet adds two convolutional layers with an important number of channels to the CNN backbone (recommended setting: $\geq$ 128). Then, it connects the prototype layer with a Sigmoid activation function. Finally, the prototypes have the same depth as the last convolutional layer (e.g., 128) for the computation of the similarity matrices with the L2 distance.
We propose to remove the two additional convolutional layers (see right block of Figure~\ref{fig:LProto}). As a consequence, the prototypes have a much smaller depth (same as CNN backbone output: 32 versus 128). This allows us to reduce the number of weights by 29.4\% compared to the original prototype block of ProtoPNet, and implies a 17.0\% reduction in CPU inference time.
Then, we replace the last activation function from the CNN backbone with a Sigmoid. As suggested in~\cite{Sandler18} and confirmed by our experiments (see Section~\ref{sec:classification}), replacing the ReLU activation function from the last layer of our CNN backbone with a Sigmoid improves the accuracy of our network. 
We further improve the accuracy of our network by adding an L2 regularization on the weights of the prototypes in order to enhance its generalization ability. Considering the limited number of prototypes, and supported by the results from our experiments, we have selected an L2 regularization over an L1 sparse solution.
Finally, \ournet\ learns a variable number of prototypes per class ($N_1$, ...,  $N_{200}$- see Figure~\ref{fig:LProto}), while the existing prototype block in ProtoPNet requires a fixed number of prototypes per class, set as hyperparameter of the model. This new feature allows \ournet\ to characterize each class with an adequate number of prototypes and better inform the end-user about the discriminative part among classes, while reducing the total number of prototypes (340 versus 400). 
Overall, our new prototype layer reduces the number of weights by 36\% and CPU inference time by 24\% compared to the original prototype block, while increasing its accuracy by 4\% - split in Section~\ref{sec:classification}
The next section details the learning procedure of \ournet.

\subsection{Network Training}
\label{sec:training}
We present \ournet\ pseudocode in Algorithm \ref{pseudocode:lexnet}. Its training procedure is composed of three stages.
First, as other prototype networks~\cite{Chen19}, the procedure updates the weights of the backbone and prototypes while keeping the weights of the last layer constant (see Stage 1.).
Next, following a predetermined number of epochs (hyperparameter $N_{SGD}$ - 20 in our setting), the procedure updates the prototypes by projecting them onto the nearest latent training patch from the same class (see Stage 2.). 
Then, in order to learn a variable number of prototypes per class (1 prototype per class at the initialization of the network to minimize the size of the model), we compute the average distance of all samples to the closest prototype of their class to evaluate how discriminative the learned prototypes are across the dataset (size of $avg\_dists: 1 \times N_{classes}$).
It is desirable to have low distances on average for all classes, which would imply that the prototypes are well representing their class. Therefore, we adopt the Kurtosis to assess the distribution of distances, i.e., a measure of the tailedness of a distribution. If the distribution is fat-tailed (kurtosis > 0 - kurtosis of the normal distribution), we add one prototype for each class in the 25-th percentile of the distribution in order to better characterize the diversity of samples in that classes.
Finally, we update the weight of the last layer only while keeping the weights of the backbone and prototypes constant for a predetermined number of epochs (hyperparameter $N_{last}$ - see Stage 3.). These three stages are repeated until reaching the global number of epochs ($N_{epochs}$).

\begin{algorithm}[!htpb]
	\caption{LEXNet}
    \scriptsize
	\label{pseudocode:lexnet}
	\begin{algorithmic}[1]
		\Initialize{$w_{back} \leftarrow$ Kaiming uniform initialization; 
                    $\forall j$: prototype $\mathbf{p}_j \leftarrow$ Uniform($[0, 1]^{H \times W \times D}$);
                    $\forall k, j: w_h^{(k, j)} \leftarrow 1$ if $\mathbf{p}_j \in \mathbf{P}_k$, $w_h^{(k, j)} \leftarrow -0.5$ if $\mathbf{p}_j \notin \mathbf{P}_k$}
		
		\For{epoch $t = 1, ..., N_{epochs}$} 
            \State \textcolor{gray}{/* Stage 1: SGD of layers before the last */}
            \For{SGD training epoch $t' = t+1, ..., t+N_{SGD}$}
                    \If{$t' < 5$} \textcolor{gray}{/* Warm-up of the backbone */}
                        \State $w_{back} \leftarrow$ update weights of the backbone;
                    \EndIf
                    \State $w_{back} \leftarrow$ update weights of the backbone;
                    \State $\mathbf{P} \leftarrow$ update weights of the prototypes;                   
            \EndFor
            $t \leftarrow t+N_{SGD}$ 

            \State \textcolor{gray}{/* Stage 2: update of the prototypes */}
            \ForEach{prototype $\mathbf{p}_j$}
                \State $\mathbf{p}_j \leftarrow$ project prototype onto the nearest latent training patch from the same class;
            \EndFor
            $dists \leftarrow$ min distance of each sample to the prototypes of its class;
            \State $avg\_dists \leftarrow$ average $dist$ per class across all samples;
            \If{kurtosis($avg\_dists$) $> 0$}
                \State $classes \leftarrow$ classes in 25-th percentile of $avg\_dists$;   
                \ForEach{k in $classes$}
                    \State $\mathbf{P}_k \leftarrow$ add 1 prototype;
                \EndFor
            \EndIf
            
            \State \textcolor{gray}{/* Stage 3: training of the last layer */}
            \For{training epoch $t'' = 1, ..., N_{last}$}
                \State $w_{h} \leftarrow$ update weights of the last layer
            \EndFor
        \EndFor
	\end{algorithmic}
\end{algorithm}

\subsection{Network Architecture}
\label{sec:architecture}
The overall architecture of \ournet\  is presented in Table~\ref{tab:ournet}.
In order to limit the size of our network, the number of LERes blocks has been determined by cross-validation on the training set. Thus, additional LERes blocks would not increase the accuracy of the network on our traffic classification dataset. A higher number of filters would not increase the accuracy either.
Concerning the explanations supporting network predictions, as presented in Section~\ref{sec:lexnet}, activation maps from the prototype layer can be upsampled to the size of the input sample to produce a heatmap that identifies the part of the input sample most similar to the learned prototype.
It has been shown in~\cite{Fauvel21} that applying upsampling processes to match the size of the input sample can affect the precision of the explanations. Therefore, we keep the feature map dimensions over the network the same as the input sample dimensions (20 $\times$ 2 - detailed in Section~\ref{sec:dataset}) using fully padded convolutions.

\begin{table}[!htpb]
	\caption{Overall architecture of \ournet.}
	\label{tab:ournet}
	\centering
	\scriptsize
	\begin{tabularx}{\linewidth}{>{\centering}m{1.5cm}>{\centering}m{1.4cm}>{\centering}m{.5cm}>{\centering}m{1.5cm}>{\centering\arraybackslash}m{1.5cm}}
		\toprule
		\textbf{Input Dims}&\textbf{Operator}&\textbf{Stride}&\textbf{Out Channels}&\textbf{Cum. Params}\\
		\midrule
		1$\times$20$\times$2 & Conv3x3+BN & 1 & 8 & 88 \\
		8$\times$20$\times$2 & LERes Block & 1 & 16 & 3,088 \\
		16$\times$20$\times$2 & LERes Block & 1 & 16 & 7,760 \\
		16$\times$20$\times$2 & LERes Block & 1 & 32 &  19,520\\
		32$\times$20$\times$2 & LERes Block & 1 & 32 & 38,080 \\
		32$\times$20$\times$2 & LProto Layer & - & 340 & 50,880 \\
		340$\times$20$\times$2 & Max Pooling & - & 340 & 50,880 \\	
		340$\times$1$\times$1 & FC & - & 200 & 118,880 \\		
		\bottomrule
	\end{tabularx}
\end{table}

\section{Evaluation Setting}
In this section, we present the methodology employed (dataset, algorithms, hyperparameters and configuration) to evaluate our approach.

\subsection{Dataset}
\label{sec:dataset}
\noindent\textbf{Commercial-Grade Dataset} Our real-world dataset~\cite{figshare} has been collected from four customer deployments in China. The dataset is composed of the TCP (e.g., HTTP) and UDP (e.g., streaming media, VoIP) traffic activity over four weeks across tens of thousands network devices. Specifically, it contains 9.7M flows/MTS belonging to 200 applications/classes (169 TCP and 31 UDP). 
For each flow/MTS, we have available some per-packet information (variables: packet size and direction).
We set the MTS length to the first 20 packets. This value reflects the relevant time windows to identify the applications and sustain line rate classification of the traffic. 
Concerning the labeling, each flow has been annotated with application names provided by a commercial-grade deep packet inspection (DPI) engine, i.e., a traditional rule-based method. Traffic encryption in China is not as present as in the Western world yet, so DPI technologies still offer fine-grained view on traffic.

Table~\ref{tab:Dataset} presents the structure of the dataset. 
We observe a class imbalance as the top 10 applications represent more than 40\% of the flows. This is characteristic of the traffic classification task and we show in Section~\ref{sec:classification} that our classifier is robust to this class imbalance.   

\begin{table}[!htpb]
	\caption{Composition of our commercial-grade dataset. Applications are presented in descending order of their popularity.}
	\label{tab:Dataset}
	\centering
	\scriptsize
	\begin{tabularx}{.93\linewidth}{>{\centering}m{1.5cm}>{\centering}m{1.5cm}>{\centering}m{1.5cm}>{\centering\arraybackslash}m{1.5cm}}
		\toprule
		\textbf{Applications}&\textbf{Flows (M)}&\textbf{Flows (\%)}&\textbf{TCP \%}\\
		\midrule
		10 & 4.1 & 42.9 & 70.0 \\
		20 & 5.7 & 59.4 & 69.5 \\
		50 & 7.8 & 80.3 & 74.5 \\
		100 & 9.0 & 92.9 & 76.2 \\
		200 & 9.7 & 100.0 & 76.6 \\
		\bottomrule
	\end{tabularx}
\end{table}

\noindent\textbf{Other Public Datasets} Moreover, to show the generalizability of our results, we have also evaluated \ournet\ on the current benchmark dataset for traffic classification (MIRAGE~\cite{aceto19mirage} - 100k flows belonging to 40 applications), the recent CESNET-TLS~\cite{Luxemburk23} (38M flows belonging to 191 applications) and on the dataset MNIST~\cite{lecun98} from computer vision (70k samples belonging to 10 classes).

\begin{table*}[!htpb]
	\caption{Accuracy with standard error and inference time of the state-of-the-art classifiers on the TCP+UDP test set.}
	\label{tab:backbone}
	\centering
	\scriptsize
	\begin{tabularx}{\linewidth}{m{2.2cm}>{\centering}m{.7cm}>{\centering}m{.7cm}>{\centering}m{.7cm}|>{\centering}m{.7cm}>{\centering}m{.7cm}>{\centering}m{.8cm}>{\centering}m{.8cm}>{\centering}m{.8cm}>{\centering}m{.8cm}>{\centering}m{.7cm}>{\centering}m{.7cm}>{\centering}m{.8cm}|>{\centering}m{.6cm}>{\centering\arraybackslash}m{.6cm}}
		\toprule
		\textbf{Metric}&\multicolumn{3}{c|}{\textbf{ResNet}}&\multirow{2}{.4cm}{\centering\textbf{1D CNN}}&\multirow{2}{.7cm}{\centering\textbf{Mobile NetV3}}&\multirow{2}{.7cm}{\centering\textbf{Shuffle NetV2}}&\multirow{2}{.8cm}{\centering\textbf{Condense NetV2}}&\multirow{2}{.8cm}{\centering\textbf{Dense Net}}&\multirow{2}{.8cm}{\centering\textbf{MLSTM FCN}}&\multirow{2}{.7cm}{\centering\textbf{XCM}}&\multirow{2}{.7cm}{\centering\textbf{Ghost Net}}&\multirow{2}{.8cm}{\centering\textbf{Efficient Net}}&\multirow{2}{.4cm}{\centering\textbf{RF}}&\multirow{2}{.4cm}{\centering\textbf{XGB}}\\
        & \textbf{Small} & \textbf{Medium} & \textbf{Large} & & & & & & & & & & & \\
		\midrule
		Accuracy (\%) & 86.7 $\pm$ 0.2 & \textbf{90.4 $\pm$ 0.1} & 89.6 $\pm$ 0.2 & 84.5 $\pm$ 0.2 & 85.7 $\pm$ 0.3 & 86.9 $\pm$ 0.3 & 87.9 $\pm$ 0.1 & 89.5 $\pm$ 0.4 & 86.5 $\pm$ 0.3 & 88.1 $\pm$ 0.2 & 88.6 $\pm$ 0.3 & 86.6 $\pm$ 0.4 & 84.4 $\pm$ 0.2 & 84.7 $\pm$ 0.2 \\	
		Inf. GPU ($\mu s$/sample) & 0.9 & 1.6 & 6.1 & \textbf{1.4} & 2.2 & 2.5 & 3.1 & 3.3 & 4.6 & 8.1 & 11.5 & 13.3 & - & - \\
		Inf. CPU ($\mu s$/sample) & 16.6 & 34.6 & 98.6 & \textbf{26.6} & 48.6 & 69.8 & 80.3 & 91.8 & 186.3 & 355 & 473.8 & 610.8 & 1.6E5 & 6.2E4 \\
		\hline
		\# Edges (k) & 138 & 303 & 1003 & 274 & 312 & 305 & 326 & 328 & 394 & 315 & 322 & 308 & 349 & 352 \\
		Mult-Adds (M) & 0.5 & 2.1 & 20.1 & 0.6 & 0.7 & 0.5 & 0.4 & 8.3 & 2.3 & 2.6 & 6.9 & 2.2 & - & - \\
		\bottomrule
	\end{tabularx}
\end{table*}

\subsection{Algorithms}
We compare our algorithm \ournet~\cite{github} to the state-of-the-art classifiers presented in Section~\ref{sec:related_work}. 
Specifically, we used the authors' implementation of CondenseNetV2~\cite{Yang21CN2}, MLSTM-FCN~\cite{Karim19}, ProtoPNet~\cite{Chen19} and XCM~\cite{Fauvel21}.
Moreover, we used the PyTorch Hub~\cite{PyTorchHub21}/ Torchvision~\cite{Torchvision21} implementations of DenseNet~\cite{Huang17}, EfficientNet~\cite{Tan19}, GhostNet~\cite{Han20}, MobileNetV3~\cite{Howard19}, ResNet~\cite{He16} and ShuffleNetV2~\cite{Ma18}.
Then, we have implemented with PyTorch in Python 3 the 1D CNN.
Finally, we used the public implementations available for the ensembles (Random Forest~\cite{scikit-learn}, XGBoost~\cite{Chen16}) and post-hoc explainability methods (Grad-CAM~\cite{Selvaraju17}, SHAP~\cite{Lundberg17}).

\subsection{Hyperparameters}
\label{sec:hyperparameters}
Adopting a conservative approach, we performed a 50\% train/50\% test split of our dataset while preserving its structure (split provided in~\cite{figshare}). We adopted the same approach for the other traffic datasets (MIRAGE, CESNET-TLS), and kept the train/test split provided for MNIST.
Hyperparameters have been set by grid search based on the best average accuracy following a stratified 5-fold cross-validation on the training set. As recommended by the authors, we let the number of LSTM cells varies in $\{8,64,128\}$ for MLSTM-FCN, and the size of the window varies in $\{20\%,40\%,60\%,80\%,100\%\}$ for XCM.
Considering the size of our input data (20 $\times$ 2), we let the hyperparameters of ProtoPNet (number and size of the prototypes) vary in $\{1,2,3,4\}$.
Similarly, we let the size of the prototypes for \ournet\ varies in $\{1,2,3,4\}$.
For the other architectures, we use the default parameters suggested by the authors.

\subsection{Configuration}
All the models have been trained with 1000 epochs, a batch size of 1024 and the following computing infrastructure: Ubuntu 20.04 operating system, GPU NVIDIA Tesla V100 with 16GB HBM2. With regard to the limited computational resources of network devices,  we also evaluate the models without GPU acceleration on an Intel Xeon Platinum 8164 CPU (2.00GHz, 71.5MB L3 Cache).

\section{Results and Discussions}
\label{sec:results}
In this section, we first present the performance results of \ournet, then, we discuss the explainability feature of our approach.

\subsection{Classification}
\label{sec:classification}
\textbf{CNN Backbone} First, in order to determine the starting architecture for the backbone of \ournet, we compare the state-of-the-art classifiers. Table~\ref{tab:backbone} presents the accuracy and inference time of the classifiers.
We observe that the model obtaining the best accuracy on our dataset is ResNet (90.4\%), while getting the lowest variability across folds (0.1\%). This model already allows us to reduce the model size by a factor of five compared to the current state-of-the-art traffic classifiers (1-2M of trainable parameters - see Section~\ref{sec:related_work} - versus 303k for ResNet). Larger or smaller ResNet models do not exhibit a higher level of accuracy (small 138k: 86.7\%, large 1M: 89.6\% - see left side of Table~\ref{tab:backbone}).
Then, Table~\ref{tab:backbone} shows all other state-of-the-art classifiers on a comparable model size for a fair comparison: $\sim$ 300k edges - trainable parameters or leafs.
ResNet exhibits the best accuracy/inference time trade-off; it obtains the second position with regard to the inference time on both GPU and CPU. 
Traditional tree-based ensemble methods - Random Forest and XGBoost - obtain an accuracy close to the 1D CNN (84.5\%), while being around a thousand times slower.
Concerning the state-of-the-art efficient CNNs, they all exhibit a higher inference time than ResNet. In particular, some efficient CNNs with really low FLOPs/Multiply-Adds (e.g., CondenseNetV2: 0.4M) compared to ResNet (2.1M) have a much higher inference time (e.g., CondenseNetV2: GPU 3.1$\mu s$/CPU 80.3$\mu s$ versus GPU 1.6$\mu s$/CPU 34.6$\mu s$). The extensive use of operations that reduce the number of FLOPS (e.g., depthwise and 1x1 convolutions) do not translate into reduction in inference time due to factors like memory access cost~\cite{Ridnik21}. Thus, our experiments show that optimizing FLOPs does not always reflect equivalently into inference time, and emphasize the interest of comparing model performance on the same model size.

Nonetheless, a model is faster than ResNet: the vanilla 1D CNN (GPU 1.4$\mu s$/CPU 26.6$\mu s$ versus GPU 1.6$\mu s$/CPU 34.6$\mu s$). The absence of residual connection in the 1D CNN can explain this lower inference time.
Therefore, we have selected ResNet as starting point for our backbone and worked to reduce its number of weights and inference time by introducing a new residual block (LERes), while maintaining its accuracy as detailed in next section.

\noindent\textbf{\ournet\ versus ProtoPNet} 
Our state-of-the-art explainable-by-design CNN baseline is ProtoPNet.
Table~\ref{tab:ournet_ablation} shows the ablation study from ProtoPNet to \ournet\ with our two contributions: a new lightweight and efficient residual block (LERes) and a lightweight prototype layer (LProto).
The performances reported correspond to the ones with the best hyperparameter configuration obtained by cross-validation on the training set: prototypes of size (1, 1) for both networks and a fixed number of 2 prototypes per class for ProtoPNet (\ournet\ learned $1.7 \pm 0.1$ prototypes per class on average).

First, ProtoPNet with ResNet as CNN backbone exhibits an accuracy of 86.2\% with 199k trainable parameters. This reduction in model size compared to ResNet (303k parameters - see Table~\ref{tab:backbone}) comes from the reduction in size of the fully connected network used for classification (ProtoPNet: 400 values as input - 2 similarity scores/prototypes per class).
Then, we can see that the replacement of the prototype block of ProtoPNet by LProto significantly reduces the number of parameters (-36\% of trainable parameters), while increasing the accuracy of the network by 4\% to reach the same accuracy as ResNet (90.4\% - see Table~\ref{tab:backbone}). As a consequence, the adoption of LProto also leads to a significant reduction in inference time (GPU: -29\%/CPU: -24\%).
Specifically, the reduction of the number of parameters and inference time mainly come from the removal of the convolutional layers, which also decreases the depth of the prototypes. And, the increase in accuracy mainly comes from the L2 regularization (+2.8\%) which improves the generalization ability of our approach.
Moreover, the introduction of a variable number of prototypes per class implies a reduction of the total number of prototypes compared to the state-of-the-art method (\ournet\: 340 -  $1.7 \pm 0.1$ per class - versus ProtoPNet: 400 - 2 prototypes per class), which lowers the number of parameters (-12k) and inference time (GPU: -0.1 $\mu s$/sample/CPU: -11.2 $\mu s$/sample), while maintaining the accuracy (see (4) in Table~\ref{tab:ournet_ablation}). 

Next, the replacement of the original residual block by LERes block allows ResNet to be faster than the vanilla 1D CNN (GPU 1.3$\mu s$/ CPU 20.3$\mu s$ versus GPU 1.4$\mu s$/CPU 26.6$\mu s$), while reducing the number of trainable parameters of its backbone by 19\% and maintaining the accuracy (99.3\% of the original ResNet).
When used in ProtoPNet, we observe in Table~\ref{tab:ournet_ablation} a comparable performance evolution as LERes block reduces the backbone parameters and CPU inference time by 19\%, while preserving the accuracy (-0.7\%).
Our LERes block integrates two new operations (generation of cost-efficient feature maps with a linear transformation, concatenation on shortcut connection), and we observe that both operations reduce the number of trainable parameters and inference time, with the first operation having twice the impact of the second one.

Overall, our explainable-by-design \ournet\ is more accurate (+4\%), lighter (-40\%) and faster (-39\%) than the current state-of-the-art explainable-by-design CNN network ProtoPNet, while not inducing a higher variability across folds (0.1\%).

\begin{table}[!htpb]
	\caption{Ablation study of \ournet. Blanks mean zero.}
	\label{tab:ournet_ablation}
	\centering
	\scriptsize
	\begin{tabularx}{\linewidth}{m{2.2cm}>{\centering}m{1.1cm}>{\centering}m{1cm}>{\centering}m{1cm}>{\centering\arraybackslash}m{1.3cm}}
		\toprule
		& \textbf{Accuracy (\%)}&\textbf{\# Params (k)}&\textbf{Inf. GPU ($\mu s$/sample)}&\textbf{Inf. CPU ($\mu s$/sample)}\\
		\midrule
		\textbf{(0) ProtoPNet} & \textbf{86.2 $\pm$ 0.1} & \textbf{199} & \textbf{5.5} & \textbf{169.7}\\
		(1):(0)+No Add. Conv & +0.2 & -59 & -1.5 & -29.3\\
		(2):(1)+Sigmoid & +1.2 &  &  & -0.6\\
		(3):(2)+L2 Regularization & +2.8 &  &  & +1\\
        (4):(3)+Var. Protos &  & -12 & -0.1 & -11.2\\
		\hline
		\textbf{(4): (0)+LProto} & \textbf{90.4 $\pm$ 0.1} & \textbf{128} & \textbf{3.9} & \textbf{129.6}\\
		Summary & +4\% & -36\% & -29\% & -24\% \\
		\hline
		(5):(4)+Linear & -0.5 & -3 & -0.1 & -11.9\\
		(6):(5)+Concatenate & -0.2 & -6 & -0.2 & -15\\
		\hline
		\textbf{\ournet: (4)+LERes} & \textbf{89.7 $\pm$ 0.1} & \textbf{119} & \textbf{3.6} & \textbf{102.7}\\
		Summary & +4\% & -40\% & -35\% & -39\% \\
		\bottomrule
	\end{tabularx}
\end{table}

\begin{figure*}[!htpb]
	\centering
	\includegraphics[width=0.99\linewidth]{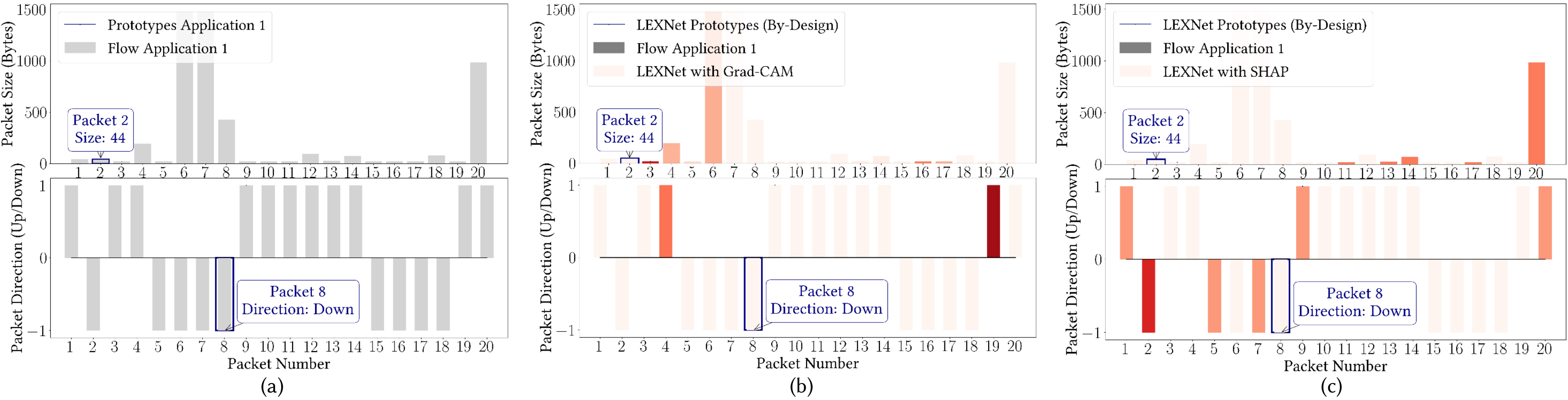}
	\caption{(a) Example of a flow from one TCP application of our dataset with the prototypes identified by \ournet\ with its explainability-by-design in blue. (b-c) Same flow with the two prototypes identified by \ournet\ with its explainability-by-design in blue versus the post hoc explainability methods Grad-CAM (b) and SHAP (c) in red.}
	\label{fig:ournet_explainability}
\end{figure*}

\begin{figure}[!htpb]
	\centering
	\includegraphics[width=.74\linewidth]{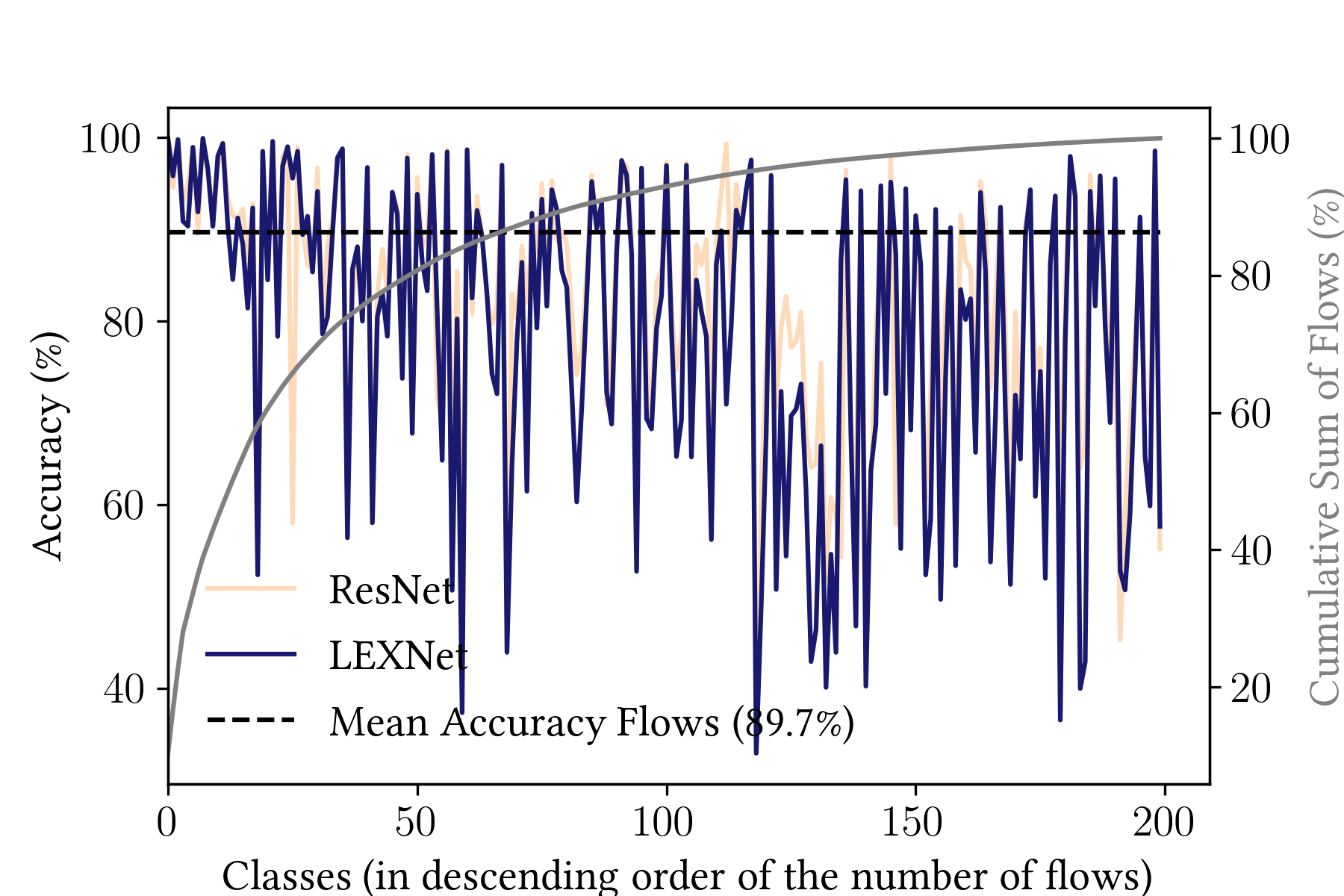}
	\caption{LEXNet and ResNet (with LERes block) accuracy per class on the TCP+UDP test set.}
	\label{fig:ournet_classimbalance}
\end{figure}

\noindent\textbf{\ournet\ Predictions}
Our results show that Internet encrypted flows can be classified with a high state-of-the-art accuracy of 89.7\% using solely the values of $1.7 \pm 0.1$ prototypes of size (1, 1) per application based on MTS containing the first 20 packets of a flow, with per-packet size and direction as variables.
More particularly, 1.7 prototypes of size (1, 1) per application on average are sufficient to classify both TCP and UDP applications with a high state-of-the-art accuracy of 87.8\% for TCP and 95.9\% for UDP (consistent with the literature~\cite{Yang21}). 
This best hyperparameter configuration (prototypes of size (1, 1)) also informs us that the information necessary to discriminate applications is often not long sequence of packet sizes or directions, but the combination of both packet sizes and directions at different places of the flow.
Then, Internet traffic is highly imbalanced, with a few applications generating most of the flows. As an example, half of the classes (100 classes) represent less than 10\% of the total number of flows in our dataset (see Table~\ref{tab:Dataset}). 
Our results show that, by maintaining the same overall high accuracy as ResNet and a comparable accuracy per class (see Figure~\ref{fig:ournet_classimbalance}), the addition of a prototype layer for explainability doesn't alter the robustness of our model to class imbalance. Among the 100 less popular classes, more than a third of them are classified with an accuracy above 75\%.

\noindent\textbf{Results on External Datasets}
To evaluate the generalizability of our results, we have also compared \ournet\ to ProtoPNet and ResNet on external datasets (see Table~\ref{tab:external_datasets}).
We observe that the results are consistent with the ones obtained on our commercial-grade traffic classification dataset. 
\ournet\ exhibits the same accuracy as ResNet while outperforming the current state-of-the-art explainable-by-design network ProtoPNet (CESNET-TLS: 97.2\% versus 94.1\%, MIRAGE: 73.6\% versus 71.4\%, MNIST: 99\% versus 97.3\%).
Moreover, \ournet\ significantly reduces the parameters and inference time compared to ProtoPNet (parameters - CESNET-TLS: 111k versus 189k, MIRAGE: 43k versus 81k, MNIST: 39k versus 70k; inference time CPU - CESNET-TLS: 92.1ms versus 146.3ms, MIRAGE: 32.8ms versus 40.9ms, MNIST: 1.1s versus 1.8s).

\begin{table}[!htpb]
	\caption{Comparison of \ournet\ and the state-of-the-art CNN with explainability-by-design on external datasets.}
	\label{tab:external_datasets}
	\centering
	\scriptsize
	\begin{tabularx}{\linewidth}{m{1.2cm}m{1.1cm}>{\centering}m{.7cm}>{\centering}m{1.1cm}>{\centering}m{1.1cm}>{\centering\arraybackslash}m{1.1cm}}
		\toprule
		\textbf{Dataset}&\textbf{Model}&\textbf{Accuracy (\%)}&\textbf{\# Params (k)}&\textbf{Inf. GPU ($\mu s$/sample)}&\textbf{Inf. CPU ($\mu s$/sample)}\\
		\midrule
		CESNET-TLS & \ournet & 97.2* & 111 & 3.4 & 92.1\\
		& ProtoPNet & 94.1 & 189 & 5.1 & 146.3\\
		MIRAGE & \ournet & 73.6* & 43 & 1.6 & 32.8\\
		 & ProtoPNet & 71.4 & 81 & 2.3 & 40.9\\	
		MNIST & \ournet & 99.0* & 39 & 16.3 & 1138\\
		& ProtoPNet & 97.3 & 70 & 25.6 & 1834\\
		\bottomrule
		\multicolumn{5}{l}{*Same accuracy as ResNet.}
	\end{tabularx}
\end{table}

\subsection{Explainability}
\label{sec:explainability}
\noindent\textbf{Illustration}
\ournet\ gives as prediction the application associated with the identified class-specific prototypes ($1.7 \pm 0.1$ prototypes on average), and as explanation the corresponding prototypes/regions of size (1, 1) on the input flow/sample as detailed in Section~\ref{sec:lexnet}.
Based on the first 20 packets of a flow from one representative TCP application (having the closest number of prototypes per class to the mean: two), Figure~\ref{fig:ournet_explainability}a illustrates this explainability by identifying in blue the two class-specific prototypes that have been used for the prediction.
The flow is represented with a bar chart for each variable and the two prototypes of size (1, 1) are highlighted by two blue rectangles identifying precisely the region of the input data that has been used for prediction. In this example, a small packet of size 44 in position 2 and a descending packet in position 8 have been detected, which are characteristic of the class containing this TCP application. 
Another application, e.g., application 2, is characterized by a large packet size of 1480 in position 8 and a descending packet in position 9. When comparing flows from different applications, this representation with prototypes allows the end-user/network expert to identify the signature of each application.

Moreover, we observe that the correlation between the number of prototypes per application of \ournet\ (ranging from 1 to 5) and the number of flows per application is negative (-0.24), which means that popular applications can be characterized by a low number of prototypes (e.g., top-20 applications - $\sim 60\%$ of the traffic - have 1 prototype per class). 
Therefore, in most cases, the explanations provided to the end-user in order to support \ournet\ predictions are easily interpretable as they only highlight one location in the flows. 
Thus, in addition to reducing the number of parameters and inference time as presented in Table~\ref{tab:ournet_ablation}, the introduction of a variable number of prototypes per class (versus a fixed number of two) provides more relevant explanations by better identifying the discriminative features of each application.

\noindent\textbf{Faithfulness}
Then, we highlight the faithfulness of \ournet\ explanations.
We compare the most important regions of size (1, 1) identified by the faithful (by definition) \ournet\ explainability-by-design - the class prototypes - to the ones from the state-of-the-art post hoc model-specific method Grad-CAM and model-agnostic method SHAP (methods used in state-of-the-art traffic classifiers - see section~\ref{sec:related_work}).
First, based on one sample from one representative TCP application, we show in Figure~\ref{fig:ournet_explainability}b-c that the post hoc explainability methods, applied on the same model \ournet, identify none of the expected regions/prototypes used by \ournet. These methods identify regions (in red) which can be far from the expected ones (e.g., Direction: Grad-CAM \#19 versus SHAP \#2 versus \ournet\ \#8).
Second, in order to quantitatively assess this difference across the dataset, considering the class-specific prototypes to be identified, we calculate the top-protos (top-1 to top-5 according to the number of prototypes per class) and top-10 accuracy in a similar fashion as the top-1 and top-5 accuracies in computer vision evaluations.
The results in Table~\ref{tab:ournet_Faithfulness} show that Grad-CAM slightly better identifies the regions of the input data that are used by \ournet\ (prototypes) compared to SHAP, but overall both post-hoc methods poorly perform (Grad-CAM: top-protos 8.2\%/top-10 38.9\%, SHAP top-protos 5.9\%/top-10 27.4\%). 
When using the 10 first predicted (1, 1) regions from the post hoc explainability methods, i.e., 25\% of the size of the input data, the expected prototypes are identified with an accuracy of less than 40\%.
Therefore, this experiment clearly emphasizes the importance of adopting explainable-by-design methods compared to post-hoc ones in order to ensure faithfulness.

\begin{table}[!htpb]
	\caption{Faithfulness evaluation on the TCP+UDP train set: ability of the state-of-the-art post hoc explainability methods to identify the prototypes used by LEXNet to predict.}
	\label{tab:ournet_Faithfulness}
	\centering
	\scriptsize
	\begin{tabularx}{\linewidth}{m{3cm}>{\centering}m{1.1cm}|>{\centering}m{1.5cm}>{\centering\arraybackslash}m{1.1cm}}
		\toprule
		LEXNet & \textbf{By-Design} & \textbf{+ Grad-CAM}&\textbf{+ SHAP}\\
		\midrule
		Top-Protos Regions Accuracy (\%) & 100.0 & 8.2 & 5.9 \\
		Top-10 Regions Accuracy (\%) & - & 38.9 & 27.4 \\
		\bottomrule
	\end{tabularx}
\end{table}

\noindent\textbf{The Cost of Explainability}
\label{sec:discussion}
Table~\ref{tab:comparison} shows the performance of \ournet\ (accuracy/size/inference) in comparison with ResNet, which can only rely on state-of-the-art post hoc explainability methods, and the current state-of-the-art explainable-by-design CNN ProtoPNet with the same backbone as \ournet.
In addition to the faithfulness, we observe that \ournet\ is around 2.5$\times$ faster than ResNet with LERes block using the most popular post-hoc explainability method for CNNs (the saliency method Grad-CAM), while maintaining the accuracy (89.7\%) and halving the model size (119k parameters).
Thus, the lightweight design of \ournet\ meets the need for fast inference that ensures line rate
classification on network devices with limited computational resources ($\sim$10k classifications/s on CPUs, versus ProtoPNet: 6k classifications/s and ResNet with Grad-CAM: 4k classifications/s), which is crucial for latency sensitive traffic.

\begin{table}[!htpb]
	\caption{Comparison of \ournet\ and the state-of-the-art CNNs with explainability-by-design and post-hoc explainability on the TCP+UDP test set.}
	\label{tab:comparison}
	\centering
	\scriptsize
	\begin{tabularx}{\linewidth}{m{2.3cm}>{\centering}m{.7cm}>{\centering}m{1.1cm}>{\centering}m{1.1cm}>{\centering\arraybackslash}m{1.1cm}}
		\toprule
		& \textbf{Accuracy (\%)}&\textbf{\# Params (k)}&\textbf{Inf. GPU ($\mu s$/sample)}&\textbf{Inf. CPU ($\mu s$/sample)}\\
		\midrule
		ResNet* + Grad-CAM & 89.7 & 294 & 9.5 & 278.6\\
		ResNet* + SHAP & 89.7 & 294 & 8.3E3 & 6.8E4\\
		ProtoPNet* & 86.2 & 199 & 5.2 & 142.8\\
		\ournet* & \textbf{89.7} & \textbf{119} & \textbf{3.6} & \textbf{102.7}\\
		\bottomrule
		\multicolumn{5}{l}{*All networks adopt LERes block to keep the same basis of comparison.}
	\end{tabularx}
\end{table}

Nonetheless, this combination obtained does not come at no cost: the explainability-by-design of LEXNet has a cost on the inference time compared to the best performing state-of-the-art CNN ResNet without explainability methods (GPU 3.6$\mu s$/CPU 102.7$\mu s$ versus GPU 1.3$\mu s$/CPU 20.3$\mu s$) and remains an open challenge. To the best of our knowledge, our study is the first to quantify the impact of explainability on the prediction performance, as well as the size and inference time of a model. Based on our experiments, around 80\% of this additional inference time is due to the L2 distance calculations in the prototype layer to generate the similarity matrices – which thus could benefit some optimization.

\section{Conclusion}
We have presented \ournet, a new lightweight, efficient and explaina- ble-by-design CNN for traffic classification which relies on a new residual block and prototype layer.
\ournet\ exhibits a significantly lower model size and inference time compared to the state-of-the-art explainable-by-design CNN ProtoPNet, while being more accurate.
Plus, \ournet\ is also lighter and faster than the best performing state-of-the-art neural network ResNet with post hoc explainability methods, while maintaining its accuracy.
Our results show that Internet encrypted flows can be classified with a high state-of-the-art accuracy using solely the values of less than two prototypes of size (1, 1) on average per application based on MTS containing the first 20 packets of a flow, with per-packet size and direction as variables.
These class prototypes detected on a flow can be given to the end-user as a simple and faithful explanation to support \ournet\ application prediction.

\ournet\ has been developed in close collaboration with our Huawei business product division. Therefore, the performance and explainability of \ournet\ meet the requirements for deployment: both for hardware computational resources (e.g., ARM CPU specifications typically used in routers - ~10k classifications/second), and for the format of explanations needed by network experts (prototypes). Concerning the usage of \ournet\ in production, in addition to traditional solutions for model monitoring (e.g., data drift detection, concept drift identification), \ournet\ explainability can be leveraged as a supplementary way to track model drift by assessing the consistency across time of the model prediction performance and the similarity to the prototypes used to support the predictions. 

In our future work, we would like to minimize the impact of the explainability feature on the inference time of \ournet\ by further optimizing the generation of the similarity matrices in the prototype layer. It would also be interesting to investigate applications in collaboration with domain experts where \ournet\ would be beneficial (e.g., mobility~\cite{Jiang19}, natural disasters early warning~\cite{Fauvel20_AAAI}, smart farming~\cite{Fauvel19}), and to further augment the capabilities of \ournet\ with such domain requirements.

\bibliographystyle{ACM-Reference-Format}
\bibliography{references}

\end{document}